\pdfoutput=1
\documentclass[11pt]{article}

\usepackage[]{acl}
\usepackage{booktabs}
\usepackage{color, soul}
\usepackage{graphicx}
\usepackage{tikz}
\usepackage{pgfplots}
\pgfplotsset{compat=newest}
\usepackage{tabu}

\usepackage{times}
\usepackage{latexsym}

\usepackage[T1]{fontenc}
\usepackage[utf8]{inputenc}

\usepackage{microtype}

\usepackage{inconsolata}
\usepackage{float}
\usepackage{multirow}
\usepackage{multicol}

\definecolor{myyellow}{HTML}{E9E29C}
\definecolor{mygreen}{HTML}{9CCB86}
\definecolor{myblue}{HTML}{009392}
\definecolor{myred}{HTML}{E88471}

\title{Toucan: Token-Aware Character Level Language Modeling}

\author{William Fleshman \\
  Johns Hopkins University\\
  \texttt{will.fleshman@jhu.edu} \\\And
  Benjamin Van Durme \\
  Johns Hopkins University \\
  \texttt{vandurme@jhu.edu} \\}

\begin{document}
\maketitle
\begin{abstract}
Character-level language models obviate the need for separately trained tokenizers, but  efficiency suffers from longer sequence lengths. Learning to combine character representations into tokens has made training these models more efficient, but they still require decoding characters individually. We propose Toucan, an augmentation to character-level models to make them \emph{``token-aware''}. Comparing our method to prior work, we demonstrate significant speed-ups in character generation without a loss in language modeling performance. We then explore differences between our learned dynamic tokenization of character sequences with popular fixed vocabulary solutions such as Byte-Pair Encoding and WordPiece, finding our approach leads to a greater amount of longer sequences tokenized as single items. Our project and code are available at \url{https://nlp.jhu.edu/nuggets/}.
\end{abstract}

\section{Introduction}

Most modern language models (LMs) are trained using the transformer architecture \cite{NIPS2017_3f5ee243} on a fixed vocabulary of tokens \cite{NEURIPS2020_1457c0d6, devlin-etal-2019-bert,touvron2023llama, penedo2023refinedweb}. Tokenizers and language models are commonly trained using separate objectives. For example, Byte-Pair-Encoding (BPE) \cite{sennrich-etal-2016-neural} selects tokens based on their frequency and not by their ability to predict the next token in a sequence. The fixed vocabulary and misaligned objectives suggest that current tokenization schemes are potentially suboptimal. 

Training transformers directly on character or byte-level sequences removes the need for tokenization, but the increased sequence length suffers from the transformer's quadratic complexity. Several variations have been developed to address the issue by pooling fixed-length contiguous character representations into smaller sets of patch representations \cite{funnel, nawrot-etal-2022-hierarchical, yu2023megabyte, tay2022charformer}. Although this can improve efficiency, it does not align with the varying length units of natural language such as words or phrases. Fixed-length pooling can also result in the same text being segmented in different ways depending on its relative position in a sequence, making the LM task on these patches more difficult.

\citet{nugget-icml} address  these issues, but rely on existing tokenization schemes. They introduce a scoring network which selects ``nuggets'' from a sequence of contextualized vectors, then pool information into those selections via transformer layers. The selected sequence of nuggets is then used to represent the text moving forward. 

\begin{figure}[t]
    \centering
    \includegraphics[width=\columnwidth]{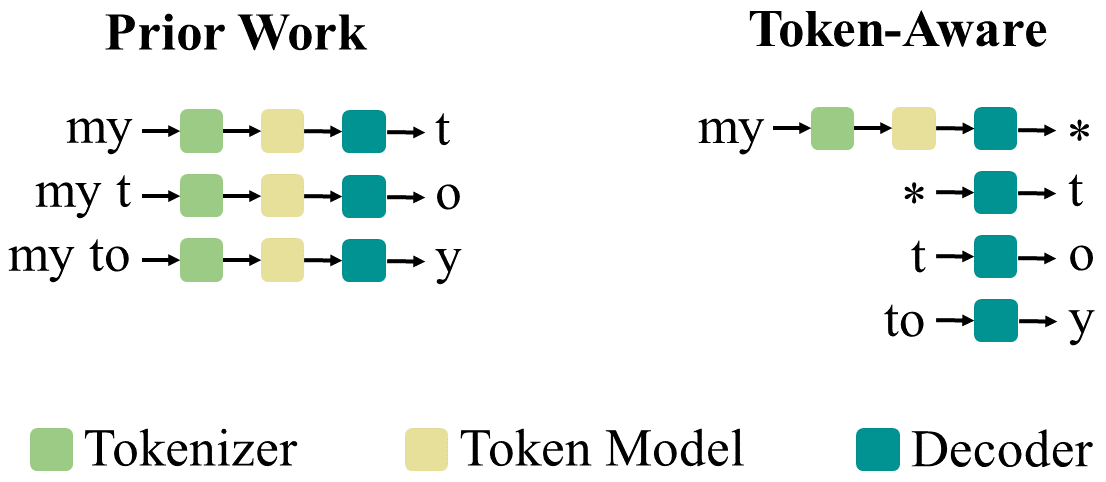}
    \caption{Token-aware generation does not require reprocessing the entire sequence at each step for every character. *: special end-of-token character.}
    \label{fig:diff}
\end{figure}

\begin{figure*}[t]
    \centering
    \includegraphics[width=.9\textwidth]{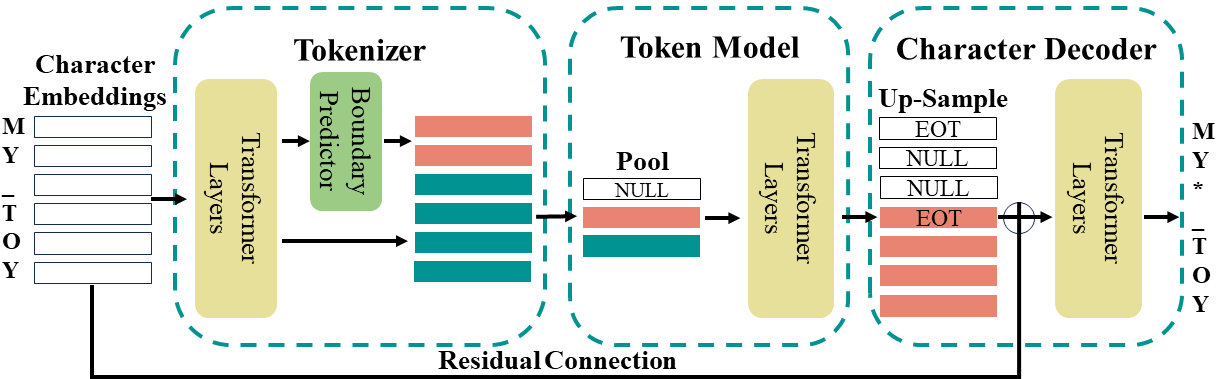}
    \caption{The architecture for Toucan, the token-aware Hourglass Transformer. End-of-token (EOT) vectors and labels (*) are inserted into the character sequences so that the decoder learns token boundaries during training. As per the original model, learned NULL vectors are used to predict the characters in the first token.}
    \label{fig:model}
\end{figure*}

\citet{nawrot-etal-2023-efficient} introduce a similar approach at the character level by modifying Hourglass Transformers \cite{nawrot-etal-2022-hierarchical} with a boundary predictor network. The predictor selects where character sequences should be segmented into tokens based on a user-defined compression prior. The segmentation is used to pool character vectors into token representations. This enables jointly training tokenization and token-level language modeling end-to-end. While the training of these models is efficient, the decoding of new text requires repeatedly passing the entire sequence through the model for every new character generated. This contrasts with token-level transformers which produce an entire token worth of characters before the sequence is reprocessed by the model. 

We therefore propose  a variant of the Hourglass Transformer with dynamic pooling augmented to become \emph{``token-aware''} in the decoding step. This approach, which we refer to as Toucan, enables decoding entire tokens using a fraction of the compute, without a loss in LM performance. An illustration of the difference is shown in Figure \ref{fig:diff}.

The contributions of this paper are as follows:

\begin{itemize}
    \item A technique for modifying character-level language models for more efficient decoding.
    \item An application of our approach to \citet{nawrot-etal-2023-efficient}'s Hourglass Transformer, resulting in over 2x faster character decoding.
    \item A comparison of popular tokenizers with those learned end-to-end with our models.
\end{itemize}

\section{Background}

\subsection{The Hourglass Transformer}

\citet{nawrot-etal-2022-hierarchical} designed the Hourglass Transformer to address challenges modeling long sequences. Specifically, they introduce contiguous fixed-width pooling at various stages of a typical transformer to shorten the effective length of the sequence being processed. They then up-sample back to the original length with a residual connection from the pre-pooled representation. 

\subsection{Dynamic Token Pooling}

\citet{nawrot-etal-2023-efficient} modified the Hourglass Transformer to perform dynamic-width pooling of character sequences. The pooled characters' representation is then processed as a token representation as in traditional transformers. Like \citet{nugget-icml}, the segmentation of these tokens is selected by a separate feed-forward network. While \citet{nawrot-etal-2023-efficient} developed several strategies for training this boundary predictor, our work focuses on their use of the gumbel-sigmoid, which allows end-to-end unsupervised learning of tokenization at a compression rate controlled with a user-defined prior. In keeping with their work, we refer to the achieved compression rate as the shortening factor (SF). Our main contribution is augmenting their architecture to significantly improve its decoding efficiency.

\subsection{Tokenizers}

We later compare the tokenization learned by our model with two popular alternatives: Byte-Pair-Encoding (BPE) \cite{sennrich-etal-2016-neural} and WordPiece \cite{wordpiece}. 

BPE first considers the unique words in a dataset. A set of learned tokens is initialized with the unique characters found among the words. The set is then iteratively expanded to a user-defined size by adding the most frequent combination of an existing token with an additional character. 

WordPiece \cite{wordpiece} is a similar tokenization algorithm popularized by its use in training BERT \cite{devlin-etal-2019-bert}. It differs from BPE in that characters that begin a token are treated as separate symbols than their counterparts internal to a token. Instead of frequency, the expansion of the token set is done based on a scoring function that prefers merging tokens that appear more frequently together than they do apart.

\section{Token-Aware Decoding}

The Toucan architecture is shown in Figure \ref{fig:model}. The three components of the architecture are derived from \citet{nawrot-etal-2023-efficient} but include changes for improving decoding efficiency. We label the three components of the architecture as the tokenizer, the token model, and the character decoder. First, the tokenizer contextualizes character embeddings and segments the characters into tokens using the boundary predictor. Character representations are pooled to form each token representation. To ensure the model is auto-regressive, the sequence of token vectors are offset using learned null vectors \cite{nawrot-etal-2023-efficient}. The token model processes the sequence of token vectors with typical transformer layers. The outputs of the token model are token-contextualized vectors which will be up-sampled and used by the character decoder to predict the characters of the next token.

Decoding a single character $x_t$ from \citet{nawrot-etal-2023-efficient}'s original model requires passing the entire sequence $x_{1:t-1}$ through all three model components, regardless of how the preceding characters had been segmented. We leverage the fact that characters segmented into the same token share the same contextualized representation after up-sampling. This representation is reused to predict each character in the next token and therefore provides an opportunity to reduce  computations. 

To increase decoding speed, we would like the decoder to generate all characters in a token without repeatedly reprocessing the entire sequence with the tokenizer and token model. To this end, we inject a learned end-of-token vector after each token in the up-sampled sequence. The labels for training the decoder are adjusted so that the last character of each token predicts an end-of-token symbol, and the injected end-of-token vector predicts the first character in the next token. We further remove the decoder's dependence on the tokenizer by moving the residual connection from the tokenizer to the embedding layer as in \citet{yu2023megabyte}. 

A trained Toucan model should  be able to generate an entire token using only the embedding layer and character decoder by sampling new characters from the decoder until the end-of-token symbol is predicted. The generated token is then appended to the sequence and processed by the entire model to begin the generation of the next token.

\section{Experimental Setup}

\subsection{Baseline and Evaluation}

We use the architecture from \citet{nawrot-etal-2023-efficient} as our baseline model and replicate their experiments on the text8 \citep{text8} and English wiki40b \citep{guo-etal-2020-wiki} datasets.\footnote{Data was gathered and preprocessed using their project repository: https://github.com/PiotrNawrot/dynamic-pooling/} We follow their exact training and evaluation procedures, model size, and hyper-parameters for both the baseline and our models. To evaluate decoding speed, we report wall-clock time while decoding characters on a single NVIDIA Quadro RTX-6000.

\subsection{Comparing Tokenizers}

We compare the tokenization learned by our model with two popular alternatives. First, we compute the number of unique tokens in our training set as reported by our learned tokenizer. We then train BPE and WordPiece models using our unique token count as the vocabulary size.\footnote{Byte-pair-encoding and WordPiece Tokenizers are trained using the Huggingface \textsc{Tokenizers} package: https://huggingface.co/docs/tokenizers/index.} We tokenize our training data with all models and provide tokenization statistics and examples in Section \ref{sec:token}. 

\section{Results}

\begin{figure}[t]
  \begin{center}
    \begin{tikzpicture}
      \begin{axis}[
          width=.8\columnwidth, 
          grid=major,
          grid style={dashed,gray!30}, 
          %xlabel=Number of Tokens Generated,
          ylabel=Seconds,
          legend style={at={(0.3, 0.9)},anchor=north}, 
        x tick label style={rotate=0}
 %         x tick label style={rotate=90,anchor=east}
        ]
        \addplot+[myblue, line width=3pt, mark size=3pt] table[col sep=comma]{data/baseline_speed.csv}; 
        \addlegendentry{baseline}
        \addplot+[myyellow, line width=3pt, mark size=3pt] table[col sep=comma]{data/eot5_speed.csv};
        \addlegendentry{toucan}
      \end{axis}
    \end{tikzpicture}
    \caption{Token generation speed as we increase the number of tokens. Both models trained using a (2,8,2) layer configuration and binomial prior of 0.2.}
    \label{fig:speed}
  \end{center}
\end{figure}
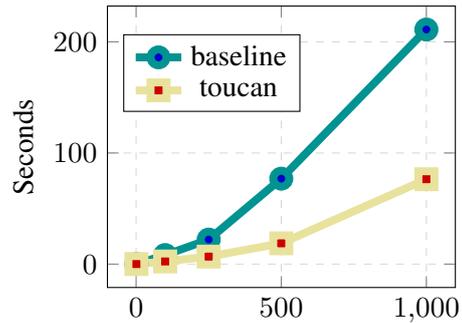

\subsection{Decoding Speed}
\label{sec:results}

As per \citet{nawrot-etal-2023-efficient}, we trained all models with a (2-8-2) layer configuration for the tokenizer, token model, and character decoder respectively. Toucan only uses the decoder for token completion, while the baseline model requires all layers for each character. The improvement in generation speed while generating an increasing number of tokens is shown in Figure \ref{fig:speed}. As expected, Toucan is significantly faster as it is using only the last two layers to produce all but the first character per token instead of the entire twelve layers required by the baseline. We verify our modifications have little impact on modeling performance and include language modeling metrics in Appendix \ref{sec:lm_results}.

\subsection{Speed Performance Tradeoff}

By reducing the binomial prior, the tokenizer is encouraged to increase its compression rate of characters. The trade off between language modeling performance, shortening factor, and generation speed is shown in Table \ref{tab:perf_speed}. The model generates characters faster with an increased shortening factor, but the language model performance suffers as a result. The models trained at the highest compression rate performed poorly; we omit them from further comparisons. 

\begin{table}[ht]
\small
\resizebox{\columnwidth}{!}{%
\begin{tabular}{cccc}
\textbf{Binomial Prior} & \multicolumn{1}{c}{\textbf{BPC}}   & \multicolumn{1}{c}{\textbf{SF}}      & \textbf{Gen@1000} 
\\
\toprule
0.05           & \multicolumn{1}{c}{1.652} & \multicolumn{1}{c}{(x24.4)} & 1.7s     \\ 
0.1            & \multicolumn{1}{c}{1.127} & \multicolumn{1}{c}{(x10.2)} & 3.9s     \\ 
0.2            & \multicolumn{1}{c}{0.997} & \multicolumn{1}{c}{(x4.9)}  & 6.1s     \\ 
\end{tabular}%
}
\caption{Bits-per-char, shortening factor, and time to generate 1k characters for varying binomial priors.}
\label{tab:perf_speed}
\end{table}

\subsection{Tokenization}
\label{sec:token}

\begin{figure}[t]
  \begin{center}
  \begin{tikzpicture}
  \begin{axis}[
      width=.9\columnwidth,
      height=4.5cm,
      grid=major,
      grid style={dashed,gray!30},
      %xlabel=Token Length,
      ylabel=Count,
      legend style={at={(0.68,0.95)},anchor=north},
      xmin=0,
      xmax=20,
      ]
    \addplot[myblue, line width=3pt] table[x, y, col sep=comma] {data/model_lengths.csv};
    \addlegendentry{toucan (x4.9)}
    
    \addplot[myyellow, line width=3pt] table[x,y, col sep=comma] {data/model10_lengths.csv};
    \addlegendentry{toucan (x10.2)}
    
    \addplot[myred, line width=3pt] table[x,y, col sep=comma] {data/bpe_lengths.csv};
    \addlegendentry{BPE/WP}

     \end{axis}
\end{tikzpicture}

\caption{Distribution of token lengths per tokenization algorithm. The plot is cut-off at token length 20, but all algorithms have thin tails extending out past this value.}
 \label{fig:distributions}

 \end{center}
\end{figure}
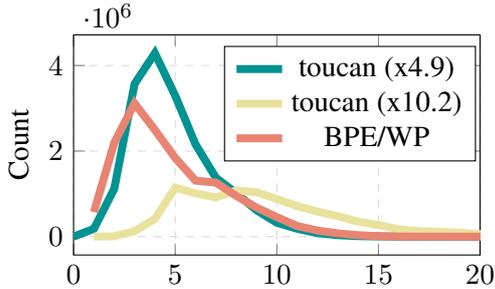

The distribution of token lengths for each model are shown in Figure \ref{fig:distributions}. The Toucan models tend to segment sequences into tokens with lengths close to their shortening factor. We plot a single distribution for BPE and WordPiece which appear nearly identical across all versions. The achieved compression rates for BPE and WordPiece were x5.8 with the smaller vocabulary and x5.9 for the larger. Because the algorithms pre-tokenize on white space, the larger vocabulary captured all unique words in the training set. We include plots for the individual models in Appendix \ref{sec:token_stats}. 

\begin{table}[ht]
\small
\resizebox{\columnwidth}{!}{%
\begin{tabular}{cccc}

\textbf{BPE} & \textbf{WP} & \textbf{Toucan (x4.9)} & \textbf{Toucan (x10.2)} \\ 
\toprule
the  & the  & \_the & \_that  \\ 
of   & of   & \_of  &\_with  \\ 
and  & and &  \_and & \_from \\
one  & one  & \_one & \_it   \\
in   & in   & \_in  & \_which \\
a   & a    & ing   & \_were  \\ 
to   & to &  \_a   & \_but      \\
zero & zero & \_to  & \_also  \\ 
nine & nine & ion   & \_eight \\
two & two  & \_zero & \_seven  \\
\end{tabular}%
}
\caption{Top ten tokens per model. Top tokens for BPE/WP remained the same for increased vocabulary. '\_': space character for the Toucan models. }
\label{tab:top-ten}
\end{table}

We show the top-10 most frequent tokens per model in Table \ref{tab:top-ten}. We observe that the Toucan (x4.9) model has similar top tokens as BPE and WordPiece but prefers segmenting suffixes more frequently than the other models. Toucan (x10.2)'s top tokens are disjoint from the other models as it tends to use those words in longer token phrases. Further comparison in Appendix \ref{sec:token_stats}.

The Toucan models tend to tokenize based on spaces, suffixes, word roots, and short phrases when using the higher shortening factor. Unlike BPE and WordPiece, the learned tokenizers can also identify tokens not seen in the training data. In Table \ref{tab:token_test} we show the tokenization of an example phrase from the test set which includes the unseen word \textit{armalite}. Further examples in Appendix \ref{sec:examples}.

\begin{table}[ht]
\resizebox{\columnwidth}{!}{%
\begin{tabular}{cc}

\textbf{Model}    & \textbf{Tokenization} \\ 
\toprule
Toucan (x4.9)  & ac:qu:is:it:ion: of: the: rifle: from: armal:ite     \\ 
Toucan (x10.2) & acquisition: of the rifle: from: armalite                     \\ \
BPE          & acquisition:of:the:rifle:from:armal:ite         \\ 
WP           & acquisition:of:the:rifle:from:arma:\#\#lite \\ 
\end{tabular}%
}
\caption{Example tokenization from  first entry in the test set. \#\# : WP marker for a token internal to words.}
\label{tab:token_test}
\end{table}

\section{Conclusion}

We proposed Toucan, a method for augmenting character-level models to generate learned tokens using a fraction of the compute compared to existing approaches. We applied Toucan to the Hourglass Transformer with dynamic token pooling and demonstrated significant speed ups in character generation without a loss in language modeling performance. We explored the differences between our learned tokenization and popular alternatives such as Byte-Pair-Encoding (BPE) and WordPiece. Our end-to-end tokenizers learn natural tokenization boundaries such as spaces, suffixes, word roots, and short phrases completely unsupervised. In contrast to Byte-Pair-Encoding and WordPiece, our tokenizers are capable of segmenting complete tokens unseen in the training data. 

\bibliography{anthology, custom}
\newpage

\appendix
\section{Language Modeling Performance}

\label{sec:lm_results}

\begin{table}[ht]
\resizebox{\columnwidth}{!}{%
\begin{tabular}{ccccccc}

            & \multicolumn{3}{c}{text8} & \multicolumn{3}{c}{wiki40b} \\ \toprule 
            & \textbf{BPC}     & \textbf{BPT}    & \textbf{SF}      & \textbf{BPC}     & \textbf{BPT}     & \textbf{SF}      \\ \midrule
baseline    & 1.195   & 5.840  & (x4.9)  & 1.115   & 5.533   & (x5.0)  \\
toucan & 0.997   & 5.911  & (x4.9)  & 0.957   & 5.699   & (x5.0)  \\ 
\end{tabular}%
}
\caption{Language model performance for the baseline and our Toucan model. Both versions were trained with a binomial prior of 0.2 encouraging a roughly (x5) shortening factor.}
\label{tab:lm_perf}
\end{table}

Our changes to the Hourglass Transformer were designed to improve decoding efficiency with minimal impact to language modeling performance. Because the Toucan model has an additional character in its vocabulary, the bits-per-character comparison is biased in our favor. Therefore, we also report bits-per-token (BPT). For an average token length $\bar{w}$ we compute bits-per-token (BPT) as
\begin{equation}
    bpt = bpc * \bar{w}.
\end{equation} 
This metric favors the baseline, because the same tokenization with Toucan will include additional bits for the end-of-token character. We show performance metrics between the baseline and our architecture in Table \ref{tab:lm_perf} and conclude that our modifications have little impact on performance.

\section{Tokenization Statistics}

\label{sec:token_stats}

 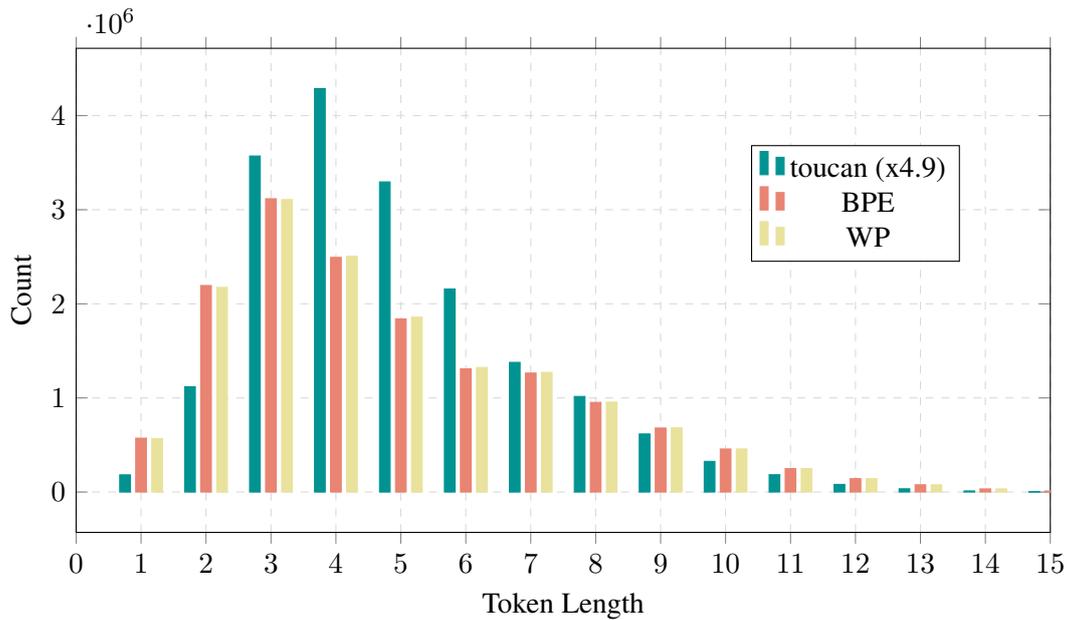
\begin{figure*}[hb]
  \begin{center}
  \begin{tikzpicture}
  \begin{axis}[
      width=0.9\textwidth, height=8cm,
      grid=major,
      grid style={dashed,gray!30},
      xlabel=Token Length,
      ylabel=Count,
      legend style={at={(0.8,0.8)},anchor=north},
      xtick=data,
      xmin=0,
      xmax=15,
      bar width=4pt,
      ybar  ]
    \addplot+[myblue, fill] table[x, y, col sep=comma] {data/model_lengths.csv};
    \addlegendentry{toucan (x4.9)}
    \addplot[myred,fill] table[x,y, col sep=comma] {data/bpe_lengths.csv};
    \addlegendentry{BPE}
    \addplot[myyellow,fill] table [x,y,col sep=comma]{data/wp_lengths.csv};
    \addlegendentry{WP}
     \end{axis}
\end{tikzpicture}
\caption{Distribution of token lengths per tokenization algorithm. BPE and WP tokenizers were trained with a vocabulary size of 192,293. The plot is cut-off at token-length 15, but all algorithms have thin tails extending out past this value.}
 \label{fig:hist_5}
 \end{center}
 \end{figure*}

\begin{figure*}[hb]
  \begin{center}
  \begin{tikzpicture}
  \begin{axis}[
      width=0.9\textwidth, height=8cm,
      grid=major,
      grid style={dashed,gray!30},
      xlabel=Token Length,
      ylabel=Count,
      legend style={at={(0.8,0.8)},anchor=north},
      xtick=data,
      xmin=0,
      xmax=20,
      bar width=4pt,
      ybar  ]
    \addplot[myblue, fill] table[x, y, col sep=comma] {data/model10_lengths.csv};
    \addlegendentry{toucan (x10.2)}
    \addplot[myred,fill] table[x,y, col sep=comma] {data/bpe_lengths_1011543.csv};
    \addlegendentry{BPE}
    \addplot[myyellow,fill] table [x,y,col sep=comma]{data/wp_lengths_1011543.csv};
    \addlegendentry{WP}
     \end{axis}
\end{tikzpicture}
\caption{Distribution of token lengths per tokenization algorithm. BPE and WP tokenizers were trained with a vocabulary size of 1,011,543. The plot is cut-off at token-length 20, but all algorithms have thin tails extending out past this value.}
 \label{fig:hist_10}

 \end{center}
 \end{figure*}
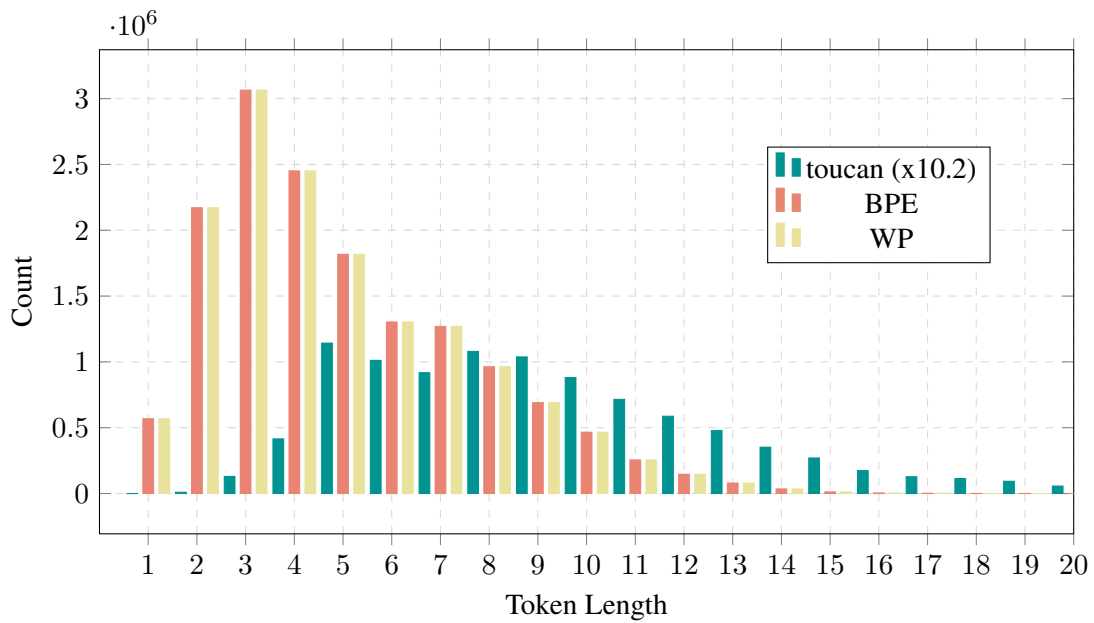

 We plot the distribution of tokens by length for each tokenization algorithm in Figures \ref{fig:hist_5} and \ref{fig:hist_10}. 
 
 Table \ref{tab:top-ten} highlighted a difference in top tokens for the Toucan (x10.2) model versus BPE and WordPiece. We report the first occurrence of their top tokens for Toucan (x10.2) in Table \ref{tab:first_occur}. The Toucan model tokenized these common words into short phrases seen frequently in the dataset.

 \begin{table}[ht]
 \small
\resizebox{\columnwidth}{!}{%
\begin{tabular}{ccc}
\
\textbf{Word} & \textbf{First Occurrence} & \textbf{Index} \\ 
\toprule
the           & the first                 & 60             \\ 
of            & of these                  & 175            \\ 
and           & and other                 & 135            \\ 
one           & one eight                 & 18             \\ 
in            & in one eight              & 107            \\
a             & a number                  & 422            \\ 
to            & to make                   & 351            \\ 
zero          & two zero zero five        & 124            \\ 
nine          & one nine eight            & 39             \\ 
two           & two zero zero five        & 124            \\ 
\end{tabular}%
}
\caption{First occurrence of top BPE/WP tokens in the Toucan (x10.2)'s top tokens.}
\label{tab:first_occur}
\end{table}

\section{Tokenization Examples}
\label{sec:examples}

We provide several example tokenizations from our test data in Tables \ref{tab:toucan_5_examples}, \ref{tab:toucan_10_examples}, \ref{tab:bpe_examples}, and \ref{tab:wp_examples}. We observe similar tokenizations from BPE and WordPiece while the Toucan (x4.9) model breaks up longer words more frequently. The Toucan (10.2) model tends to group whole words and short phrases as tokens.

\begin{table}[ht]
\resizebox{\columnwidth}{!}{%
\begin{tabular}{c}

\textbf{Toucan (x4.9)}                                        \\ \toprule
his: career: desp:ite: announc:ing: plans: to: ret:ire        \\ 
eleven: straight: commerc:ial: disappointments                \\ 
they: are: temperature: pres:sure: water: vapor               \\ 
writes: on: the: mod:if:icat:ion: of: clouds                  \\ 
includes: eukaryotes: with: a: nucleus: such: as: fung:i      \\ \midrule
capac:ity: of: hard: drives: was: measured: in: megabytes     \\ 
accused: of: irregular:it:ies: in: invest:igat:ing            \\ 
geolog:ists: to: refer: to: an: extraterrestr:ial: mesa       \\ 
in: engl:ish: poetry: feet: are: determ:ined: by: emphas:is   \\ 
mistakes: could: be: corrected: by: apply:ing: correct:ion    \\ \midrule
pol:ish: parl:iament: in: september: one: nine: nine: seven   \\ 
nasal: lateral:ity: is: the: release: of: airflow             \\ 
a: motherboard: also: known: as: a: mainboard: log:ic: board  \\ 
tombs: insp:ire: the: ant:i: arch:itectural                   \\ 
salt: cellar: of: gold: and: ebony: in: one: five: four: zero \\ \midrule
is: called: a: capac:it:ive: manometer: vacuum: gauge         \\
a: relat:ively: late: development: reconstruct:ion            \\
microwaves: at: a: frequency: of: two: four: gigahertz        \\ 
antony: octav:ian: became: uncontested: ruler: of: rome       \\ 
morphogenes:is: from: the: greek: morph: shape: and: genes:is \\ 
\end{tabular}%
}
\caption{Tokenization of phrases from the test data using Toucan (x4.9).}
\label{tab:toucan_5_examples}
\end{table}

\begin{table}[ht]
\resizebox{\columnwidth}{!}{%
\begin{tabular}{c}

\textbf{Toucan (x10.2)}                                   \\ \toprule
his career: despite: announcing: plans: to retire         \\ 
eleven: straight: commercial: disappointments             \\ 
they: are temperature: pressure: water: vapor             \\ 
writes: on the modification: of clouds                    \\ 
includes: eukaryotes: with: a nucleus: such: as fungi     \\ \midrule
capacity: of hard: drives: was measured: in megabytes     \\ 
accused: of irregularities: in investigating              \\
geologists: to refer: to an extraterrestrial: mesa        \\ 
in english: poetry: feet: are determined: by emphasis     \\ 
mistakes: could: be corrected: by applying: correction    \\ \midrule
polish: parliament: in september: one nine nine seven     \\ 
nasal: laterality: is the release: of airflow             \\
a motherboard: also: known: as a mainboard: logic: board  \\
tombs: inspire: the anti: architectural                   \\ 
salt: cellar: of gold: and ebony: in one five four zero   \\ \midrule
is called: a capacitive: manometer: vacuum: gauge         \\ 
a relatively: late: development: reconstruction           \\ 
microwaves: at a frequency: of two four gigahertz         \\ 
antony: octavian: became: uncontested: ruler: of rome     \\ 
morphogenesis: from: the greek: morph: shape: and genesis \\ 
\end{tabular}%
}
\caption{Tokenization of phrases from the test data using Toucan (x10.2).}
\label{tab:toucan_10_examples}
\end{table}

\begin{table}[ht]
\resizebox{\columnwidth}{!}{%
\begin{tabular}{c}

\textbf{Byte-Pair-Encoding}                          \\ \toprule
his:career:despite:announcing:plans:to:retire        \\ 
eleven:straight:commercial:disappointments           \\ 
they:are:temperature:pressure:water:vapor            \\ 
writes:on:the:modification:of:clouds                 \\ 
includes:eukaryotes:with:a:nucleus:such:as:fungi     \\ \midrule
capacity:of:hard:drives:was:measured:in:megabytes    \\ 
accused:of:irregularities:in:investigating           \\ 
geologists:to:refer:to:an:extraterrestrial:mesa      \\ 
in:english:poetry:feet:are:determined:by:emphasis    \\ 
mistakes:could:be:corrected:by:applying:correction   \\ \midrule
polish:parliament:in:september:one:nine:nine:seven   \\ \
nasal:later:ality:is:the:release:of:airflow          \\ \
a:motherboard:also:known:as:a:mainboard:logic:board  \\ \
tombs:inspire:the:anti:architectural                 \\ \
salt:cellar:of:gold:and:ebony:in:one:five:four:zero  \\ \midrule
is:called:a:capacitive:man:ometer:vacuum:gauge       \\ 
a:relatively:late:development:reconstruction         \\ 
microwaves:at:a:frequency:of:two:four:gigahertz      \\ 
antony:octavian:became:uncontested:ruler:of:rome     \\ 
morphogenesis:from:the:greek:morph:shape:and:genesis \\ 
\end{tabular}%
}
\caption{Tokenization of phrases from the test data using Byte-Pair-Encoding.}
\label{tab:bpe_examples}
\end{table}

\begin{table}[ht]
\resizebox{\columnwidth}{!}{%
\begin{tabular}{c}
\textbf{WordPiece}                                   \\ \toprule
his:career:despite:announcing:plans:to:retire        \\ 
eleven:straight:commercial:disappointments           \\ 
they:are:temperature:pressure:water:vapor            \\ 
writes:on:the:modification:of:clouds                 \\ 
includes:eukaryotes:with:a:nucleus:such:as:fungi     \\ \midrule
capacity:of:hard:drives:was:measured:in:megabytes    \\ 
accused:of:irregularities:in:investigating           \\ 
geologists:to:refer:to:an:extraterrestrial:mesa      \\ 
in:english:poetry:feet:are:determined:by:emphasis    \\ 
mistakes:could:be:corrected:by:applying:correction   \\ \midrule
polish:parliament:in:september:one:nine:nine:seven   \\ 
nasal:lateral:\#\#ity:is:the:release:of:airflow      \\ 
a:motherboard:also:known:as:a:mainboard:logic:board  \\ 
tombs:inspire:the:anti:architectural                 \\ 
salt:cellar:of:gold:and:ebony:in:one:five:four:zero  \\ \midrule
is:called:a:capacitive:mano:\#\#meter:vacuum:gauge   \\ 
a:relatively:late:development:reconstruction         \\ 
microwaves:at:a:frequency:of:two:four:gigahertz      \\ 
antony:octavian:became:uncontested:ruler:of:rome     \\ 
morphogenesis:from:the:greek:morph:shape:and:genesis \\ 
\end{tabular}%
}
\caption{Tokenization of phrases from the test data using WordPiece.}
\label{tab:wp_examples}
\end{table}

\end{document}